\documentclass[conference]{IEEEtran}
\IEEEoverridecommandlockouts

\usepackage{cite}
\usepackage{amsmath,amssymb,amsfonts}
\usepackage{algorithmic}
\usepackage{graphicx}
\usepackage{textcomp}
\usepackage{xcolor}
\usepackage{graphicx}
\usepackage{subcaption}
\usepackage{booktabs}
\usepackage{array} 
\usepackage{tabularx}
\usepackage{tikz}
\usepackage{url}
\usetikzlibrary{arrows.meta, positioning, shapes.geometric}
\def\BibTeX{{\rm B\kern-.05em{\sc i\kern-.025em b}\kern-.08em
    T\kern-.1667em\lower.7ex\hbox{E}\kern-.125emX}}
\begin{document}

\title{Quantifying Holistic Review: A Multi-Modal Approach to College Admissions Prediction\\}

\author{\IEEEauthorblockN{1\textsuperscript{st} Jun-Wei Zeng*}
\IEEEauthorblockA{\textit{Vanke Meisha Academy} \\
Shenzhen, China \\
zengjunwei@stu.vma.edu.cn\\
0009-0006-1920-7118}
\and
\IEEEauthorblockN{2\textsuperscript{nd} Jerry Shen}
\IEEEauthorblockA{\textit{Vanke Meisha Academy} \\
Shenzhen, China \\
shenruiyi@stu.vma.edu.cn\\
}
}

\maketitle
\vspace{200cm}
\begin{abstract}
This paper introduces the Comprehensive Applicant Profile Score (CAPS), a novel multi-modal framework designed to quantitatively model and interpret holistic college admissions evaluations. CAPS decomposes applicant profiles into three interpretable components: academic performance (Standardized Academic Score, SAS), essay quality (Essay Quality Index, EQI), and extracurricular engagement (Extracurricular Impact Score, EIS). Leveraging transformer-based semantic embeddings, LLM scoring, and XGBoost regression. CAPS provides transparent and explainable evaluations aligned with human judgment. Experiments on a synthetic but realistic dataset demonstrate strong performance, achieving an EQI prediction $R^2$ of 0.80, classification accuracy over 75\%, and a macro F1 score of 0.69, and weighted F1 score of 0.74. CAPS addresses key limitations in traditional holistic review—particularly the opacity, inconsistency, and anxiety faced by applicants—thus paving the way for more equitable and data-informed admissions practices.
\end{abstract}

\begin{IEEEkeywords}
Interpretable machine learning, XGBoost regression, Essay evaluation, Natural language processing, Principal Component Analysis, score fusion
\end{IEEEkeywords}

\section{Introduction}
    Nowadays, in the highly competitive environment of US college applications,
especially among top-tier universities, the evaluation process considers not only quantitative standardized scores and academic standards, such as the SAT and grade point average (GPA), but also qualitative factors like extracurricular activities, personal essays, and the background of the applicant, such as different extracurricular activities [2]. While the holistic review introduced a more human-centered approach to the admission process, its lack of transparency and subjectivity has raised different concerns among applicants and educators [9].

    With more than 2 million students applying to universities in the US, the acceptance rates at elite institutions dropping below 5\% [1], the risk that the applicants take to apply for their dream schools has never been higher. Traditional holistic review, while comprehensive, often leaves applicants with uncertainties about their competitive standing in multiple dimensions: academic, extracurricular and essays. Creating anxiety and information asymmetry in the application process [9].

    To address these limitations, we propose \textbf{Comprehensive Applicant Profile Score (CAPS)}, a multi-modal framework that quantifies an applicant's holistic profile using a combination of academic, essay, and extracurricular. CAPS consists of three different interpretable elements: the \textbf{Standardized Academic Score (SAS)}, the \textbf{Essay Quality Index (EQI)}, and the \textbf{Extracurricular Impact Score (EIS)}. Each module is designed to capture all the distinct aspects of the holistic review process. The final CAPS score will be computed through a fusion of weightings, allowing consistency and personalization.

\begin{table}[ht]
\centering
\caption{Comparison between Traditional and the CAPS Framework}
\resizebox{\linewidth}{!}{%
\begin{tabular}{|l|l|l|}
\hline
\textbf{Feature} & \textbf{Traditional Framework} & \textbf{CAPS Framework} \\
\hline
Data Used & Only GPA, SAT & GPA + Essays + Extracurriculars (EC) \\
\hline
Decision Basis & Opaque Heuristics & Transparent ML-based Model \\
\hline
Interpretability & Uninterpretable & SHAP + LLM-based Explanations \\
\hline
Evaluation Process & Manual Review & AI-Aided Modular Scoring \\
\hline
\end{tabular}%
}
\label{tab:comparison_caps}
\end{table}

\section{Related Work}

\subsection{College Admissions Prediction and Holistic Review}
Previous work in university admissions prediction mainly focused on structured academic metrics such as GPA and standardized test scores. Recent studies have explored machine learning (ML) support for holistic review [3,6], recognizing that traditional academic stats alone are insufficient for comprehensive evaluation. One notable study trained an admission-prediction model that replaces standardized tests by learning from historical data, achieving similar performance while improving fairness across different ethnicities [5]. This shows the application of Machine learning in education.

Recent studies have also addressed the challenge of validating admission committee decisions for undergraduate admissions, noting that traditional review processes are overwhelmed with large volumes of applicant data and remain susceptible to human bias [3]. This study employs deep learning (DL) approaches to verify quantitative assessments made by application reviewers, but many focus on validation but not holistic review prediction of comprehension evaluation.

Contemporary work has shown that while ML models can partially compensate for the removal of protected attributes (e.g., race, gender), such models still fall short in ensuring diverse admission outcomes [7].

Most existing approaches treat different components separately rather than as part of a unified, comprehensive holistic framework that can provide component-level interpretability [10].

\subsection{Multi-modal Assessment in Educational Contexts}
The integration of textual and structured data in education assessment has gained significant attention. Automated essay scoring (AES) [8] has evolved from traditional handcrafted features to neural approaches- particularly those based on Long Short-Term Memory (LSTM) networks can outperform traditional baselines without manual feature engineering [8]. Transformer-based models have further advanced and enhanced the field, demonstrating superior performance over bag-of-words and logistic regression baselines, especially for tasks requiring contextual understanding, such as politeness detection and emotional expression in written responses [11].

In the broader context of academic success prediction, researchers have modelled GPA outcomes using psychological, sociological, and academic factors, often finding that random forest regression yields the most accurate predictions [12]. However, these approaches typically focus on individual outcomes but not the holistic admissions decision-making process.

\subsection{Explainable AI in Admissions}

Given the high-stakes nature of college admission decisions, interpretability and transparency have become crucial and paramount. SHapley Additive exPlanations (SHAP) have emerged as a powerful tool for attributing feature importance in complex models [13]. The emergence of SHAP provides a compelling basis for explainable admissions modeling, which is both essential for institutions and applicants.

While recent studies have developed sophisticated college admission prediction systems, many of these models lack the comprehensive interpretability needed for such critical decisions.

\subsection{Integration of Textual Data in Holistic Review}

In recent studies, researchers have explored integrating textual data - such as personal statements, application essays and recommendation letters into holistic review systems [2]. However, most existing work focuses on isolated assessment of essay quality [8] rather than incorporating essays with different parts to form a comprehensive review process.

The challenge does not only lie in accurately accessing individual components, but also in determining the weights of different components in an essay, and even the weightings of different components of a holistic review.

\subsection{Gap in Current Research}

Despite these former researches and advancements, existing approaches suffer from several key limitations. First of all, most systems focus on individual components (essays evaluation, academics scores, or activities banding) rather than providing integrated, holistic evaluation. Secondly, current ML approaches often lack the explainability and instant feedback for both applicants and admissions committees. Finally, the quantification of "holistic review" varies significantly across institutions—remains largely unexplored and lacks transparency.

\subsection{Our Contribution}

In contrast to prior works, our study aims to build a fully explainable, modular framework that addresses the limitations of former works. While recent studies have focused on validating the admission process through interpretable deep learning (DL) approaches[3], our CAPS system provides a comprehensive three-module architecture: essay analysis (EQI), academic indicators (SAS), and extracurricular evaluation (EIS).

Specifically, our approach distinguishes itself through the following contributions:
\begin{enumerate}
    \item \textbf{Granular component-level scoring}: CAPS produces interpretable decisions, enabling fine-grained diagnostics and analysis.
    \item \textbf{Advanced multi-modal fusion}: We integrate LLM-based assessments, NLP embeddings, and traditional approaches to model holistic review.
\end{enumerate}

\begin{figure}[ht]
    \centering
    \includegraphics[width=0.7\linewidth]{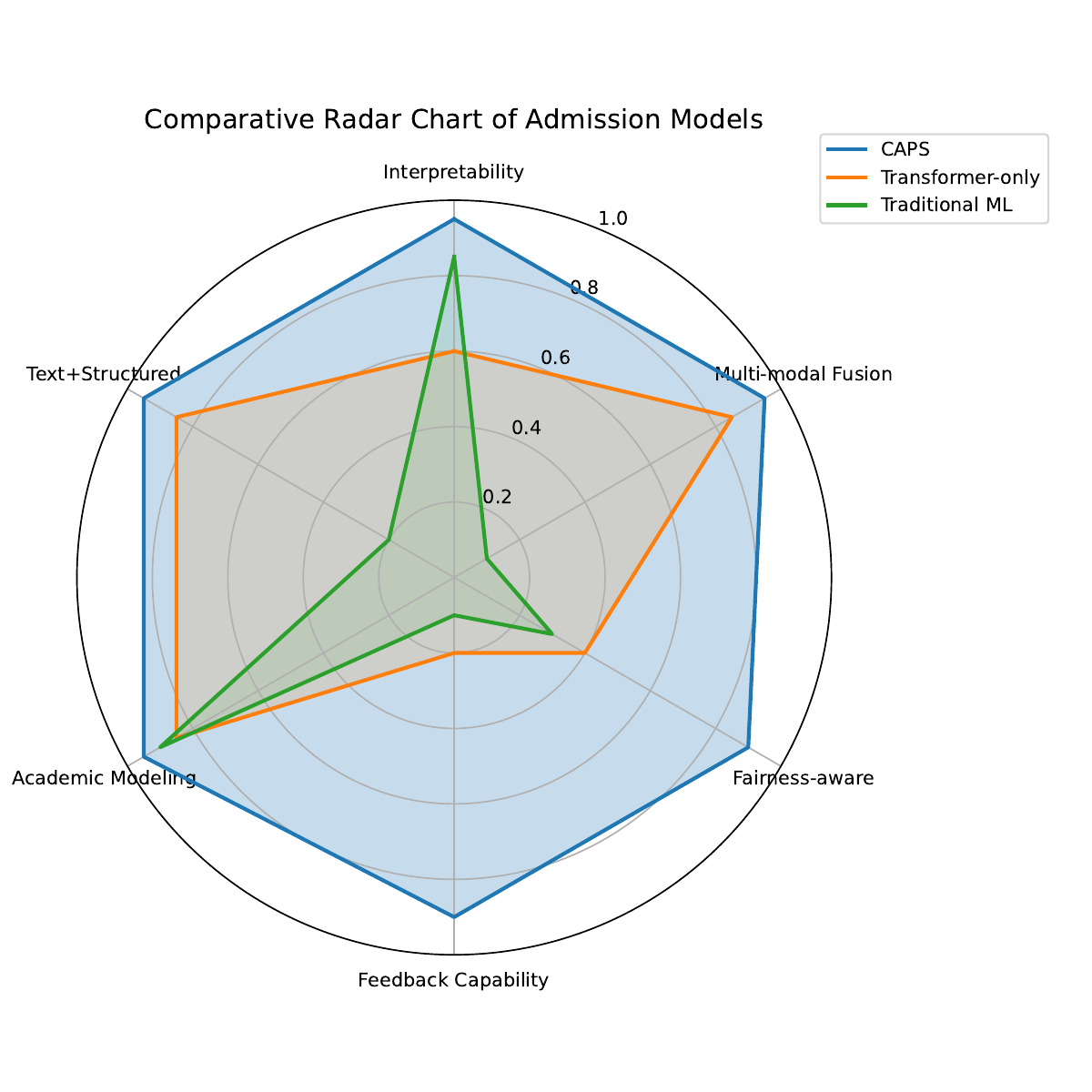}
    \caption{Radar comparison of CAPS, Transformer-only, and traditional ML models across six admission-related dimensions.}
    \label{fig:pca_academic}
\end{figure}

\section{Methodology}
\subsection{Overview of CAPS Framework}

The \textbf{Comprehensive Applicant Profile Score (CAPS)} framework is a modular, interpretable system designed to simulate holistic college admissions decisions. It integrates academic metrics normalization, Natural Language Processing (NLP) -derived essay embeddings, and GPT-evaluated qualitative judgments across three core components:

\begin{itemize}
    \item \textbf{SAS (Standardized Academic Score)} quantifies academic performance.
    
    \item \textbf{EQI (Essay Quality Index)} quantifies essays.
    
    \item \textbf{EIS (Extracurricular Impact Score)} quantifies extracurriculars.
\end{itemize}

To compute the final CAPS, outputs from the three modules are combined using the weightings computed by:
\begin{itemize}
    \item \textit{\textbf{Logistic regression coefficients}}, capturing interpretable linear relationships.
    \item \textit{\textbf{XGBoost-learned feature importances}}, capturing nonlinear patterns in admission outcomes.
    \item \textit{\textbf{Expert defined fixed weights}}, based on admissions heuristics.
\end{itemize}

These weights are fused using tunable parameters $(\alpha, \beta, \gamma)$ and normalized to produce a final score. 

The final CAPS shows the strength of the applicants' background holistically, offering transparency and interpretability across the decision pipeline.
\begin{figure}[ht]
    \centering
    \includegraphics[width=0.9\linewidth]{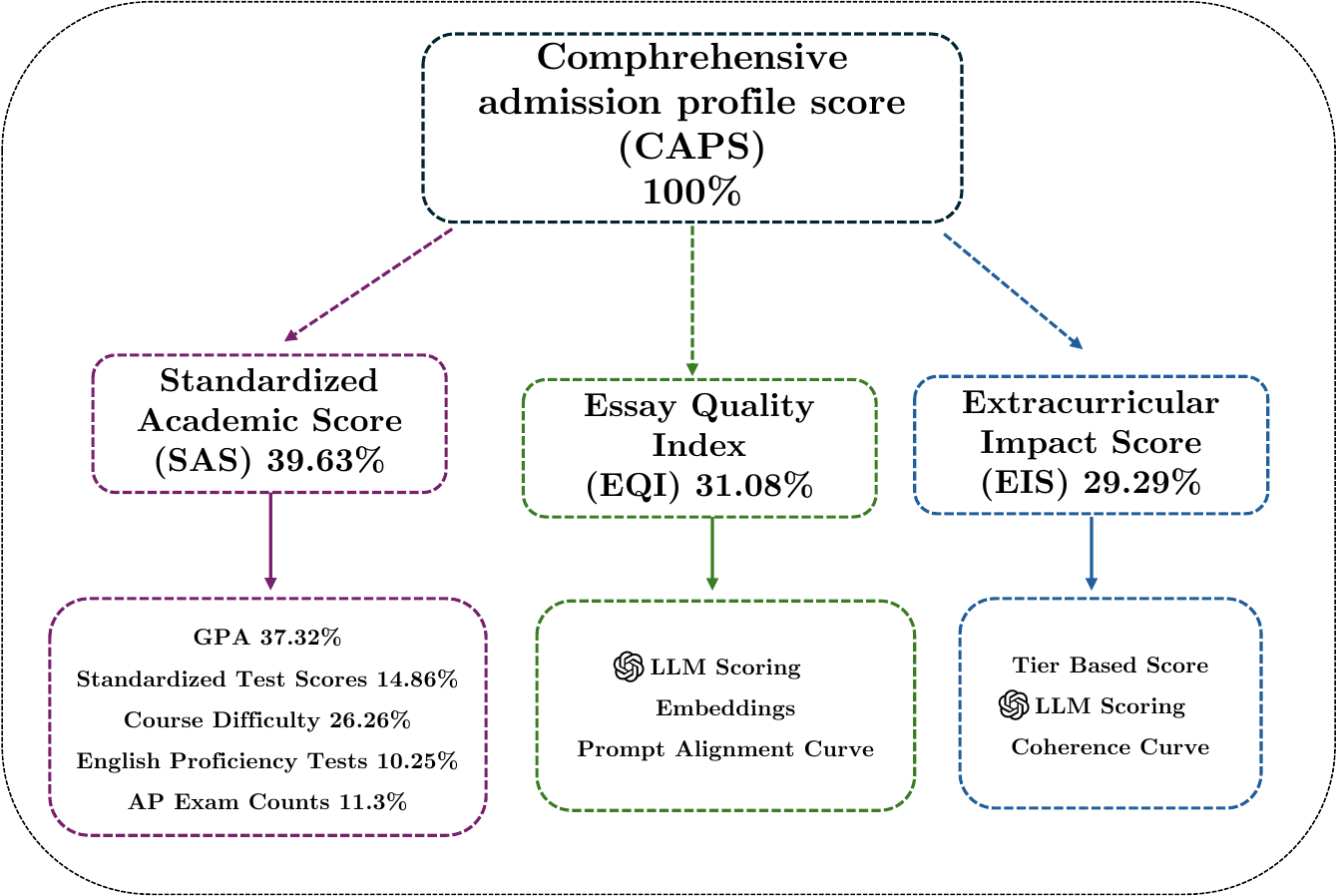}
    \caption{Framework of the CAPS }
    \label{fig:Workflow}
\end{figure}

\subsection{SAS Module}
The \textbf{Standardized Academic Score (SAS)} quantifies an applicant's academic ability based on multiple metrics: \textbf{GPA}, \textbf{standardized tests scores (SAT/ACT)}, \textbf{English proficiency test scores (TOEFL/IELTS)[16]}, the number of \textbf{Advanced Placement (AP) exams scored at 5}, and the \textbf{rigor of high school coursework (Course Difficulty)}.

These input features are standardized using the Z-score normalization method via a StandardScaler, ensuring each feature contributes equally to the final analysis without bias toward larger numeric scales.
To standardize each input feature, we apply Z-score normalization using the formula:

\begin{samepage}
\begin{equation}
z_i = \frac{x_i - \mu}{\sigma}
\end{equation}
where \( x_i \) is the original feature value, \( \mu \) is the mean of the feature across the training set, and \( \sigma \) is the standard deviation.
\end{samepage}
\subsubsection{Step 1: Hybrid Weighting Strategy}
To determine the relative weightings of each academic component in the SAS module, we employed two approaches: expert-defined weight and Principal Component Analysis (PCA) weights to achieve the most authentic weighting.

Firstly, to capture underlying relationships and reduce feature redundancy, Principal Component Analysis (PCA) is employed, reducing dimensionality from five features down to two principal components. These PCA-derived components are weighted  $(\alpha_{\text{PCA}} = 1.0, \beta_{\text{PCA}} = 0.5)$ to compute a directional PCA score, capturing the maximum variance across academic profiles.

To extract a directional importance vector from the top two principal components, we first standardize the feature matrix $X \in \mathbb{R}^{n \times d}$ using Z-score normalization:

\begin{equation}
X_{\text{scaled}} = \text{StandardScaler}(X)
\end{equation}

We then apply Principal Component Analysis (PCA) to obtain the first two components:

\begin{equation}
\text{PC}_1, \text{PC}_2 = \text{PCA}(X_{\text{scaled}})
\end{equation}

Next, we compute a linear combination of the top two principal components with tunable hyperparameters $\alpha_{\text{PCA}}$ and $\beta_{\text{PCA}}$ to obtain the directional PCA-based importance vector $w_{\text{PCA}}^{\text{raw}}$:

\begin{equation}
w_{\text{PCA}}^{\text{raw}} = -(\alpha_{\text{PCA}} \cdot \text{PC}_1 + \beta_{\text{PCA}} \cdot \text{PC}_2)
\end{equation}

To ensure the resulting vector can be interpreted as a weight distribution, we normalize it to unit sum:

\begin{equation}
w_{\text{PCA}} = \frac{w_{\text{PCA}}^{\text{raw}}}{\sum_i w_{\text{PCA}, i}^{\text{raw}}}
\end{equation}

In our implementation, we set $\alpha_{\text{PCA}} = 1.0$ and $\beta_{\text{PCA}} = 0.5$, empirically balancing between dominant and secondary variance directions.

To ensure both data-driven and human-expert-driven insights, a hybrid weighting strategy is introduced:
\begin{samepage}
\begin{itemize}
    \item \textbf{PCA-derived weights}: The raw PCA weights from principal components are first computed, adjusted, and then normalized.
    
    \item \textbf{Manual expert-defined weights}: Domain experts assign intuitive importance values based on historical admission data and educational insights [15]:
\end{itemize}
\end{samepage}
\vspace{1em}

    To combine data-driven and expert-defined knowledge, we compute the final fused weight for each academic feature using the following linear combination:
\begin{equation}
w_{\text{fused}} = \alpha_{\text{fusion}} \cdot w_{\text{PCA}} + (1 - \alpha_{\text{fusion}}) \cdot w_{\text{manual}}
\end{equation}

where $w_{\text{fused}}$ is the final fused weight, $w_{\text{PCA}}$ is the PCA-derived weight, $w_{\text{manual}}$ is the expert-defined weight, and $\alpha_{\text{fusion}} \in [0,1]$ controls the trade-off between data-driven and domain-informed weighting.

By projecting the standardized features onto two principal components, PCA highlights the underlying academic strength patterns across the applicant pool, allowing data-driven extraction of feature importance weights based on variance contribution.

In our implementation, we set \(\alpha_{\text{fusion}} = 0.1\), maintaining the traditional college admission officer's determined weightings while still incorporating the statistical signal from PCA.

\begin{table}[ht]
\centering
\caption{Comparison of Feature Weights in SAS: Manual vs. PCA-derived vs. Fused}
\label{tab:sas_weights}
\renewcommand{\arraystretch}{1.2}
\begin{tabularx}{\linewidth}{l >{\centering\arraybackslash}X >{\centering\arraybackslash}X >{\centering\arraybackslash}X}
\toprule
\textbf{Feature} & \textbf{Manual Weight} & \textbf{PCA-derived Weight} & \textbf{Fused Weight} \\
\midrule
GPA                & 0.4000 & 0.1325 & 0.3732 \\
SAT                & 0.1500 & 0.1362 & 0.1486 \\
TOEFL              & 0.1000 & 0.1249 & 0.1025 \\
AP\_5\_Count        & 0.1000 & 0.2300 & 0.1130 \\
Course\_Difficulty & 0.2500 & 0.3765 & 0.2626 \\
\bottomrule
\end{tabularx}
\end{table}
\subsubsection{Step 2: Score Computation}
    The raw Standardized Academic Score (\textbf{SAS\textsubscript{raw}}) can then be computed as the dot product between the standardized academic feature vector and the fused feature weight vector:

\begin{equation}
\text{SAS}_{\text{raw}}^{(i)} = \sum_{j=1}^{d} z_j^{(i)} \cdot w_j^{\text{fused}} = \mathbf{z}^{(i)} \cdot \mathbf{w}_{\text{fused}}
\end{equation}

Here, $\mathbf{z}^{(i)}$ denotes the $d$-dimensional z-score standardized academic feature vector for the $i$-th applicant, and $\mathbf{w}_{\text{fused}}$ is the fused weight vector combining PCA-derived and expert-defined weights. This formulation provides a linear aggregation of academic indicators, resulting in a continuous and interpretable scalar score.

The resulting vector $\text{SAS}_{\text{raw}}$ is then transformed via a \textit{softmax} function to emphasize relative performance among applicants:

\begin{equation}
\text{SAS}_{\text{softmax}}^{(i)} = \frac{\exp(\text{SAS}_{\text{raw}}^{(i)})}{\sum_{k=1}^{n} \exp(\text{SAS}_{\text{raw}}^{(k)})}
\end{equation}

Finally, we apply a sigmoid transformation scaled as follows:

\begin{equation}
\text{SAS}_{\text{scaled}} = 100 \times \sigma((\text{SAS}_{\text{softmax}} - 1.5) \times 2.5)
\end{equation}

This ensures scores smoothly map to a 0–100 scale, facilitating intuitive interpretations.
    
\subsection{EQI Module}
The Essay Quality Index (EQI) module provides a sophisticated NLP-based evaluation of applicants’ essays, measuring their semantic representation.  It leverages lightweight transformer-based embeddings (\texttt{all-MiniLM-L6-v2}) [17] to encode essay meaning, and employs large language models (LLMs) such as GPT4o [18] for dimension-specific scoring in content, language, and structure. Together with an XGBoost regression model, EQI ensures both scoring accuracy and interpretability in holistic essay assessment.

\subsubsection{\textbf{Essay Scoring Pipeline}}
\vspace{0.5em}
\paragraph{Step 1: LLM-based Essay Scoring}  
To obtain structured scores for \textbf{Content}, \textbf{Language}, and \textbf{Structure}, we provide GPT-4o with the following prompt:
\begin{quote}
\small
\begin{flushleft}
\texttt{You are an experienced admissions officer at a top U.S. university. Please evaluate the following college essay and assign a score from 1 to 5, just as you would during application review: }
\vspace{0.5em}
\texttt{1. Content: Is the theme original and does it demonstrate depth of thought?} \\
\vspace{0.5em}
\texttt{2. Language: Is the word choice precise and natural?} \\
\vspace{0.5em}
\texttt{3. Structure: Does it have a compelling introduction, smooth transitions, and a clear conclusion?}
\end{flushleft}
\end{quote}
These scores provide a structured numeric representation of essay quality.

\paragraph{Step 2: NLP Semantic Embeddings}  
Essays are encoded into 384-dimensional semantic vectors using the \texttt{all-MiniLM-L6-v2} model [17]. These embeddings capture deep contextual semantics beyond surface-level lexical similarity. Essays are encoded through the pipeline see Fig.~\ref{fig:minilm_embedding}).

\begin{itemize}
    \item \textbf{Tokenization}: The essay is split into subword tokens using a WordPiece tokenizer.
    
    \item \textbf{Embedding Layer}: Each token ID is mapped to a 384-dimensional vector with positional encoding.
    
    \item \textbf{Transformer Encoder}: The embeddings are passed through 6 Transformer layers with self-attention and feedforward networks to capture contextual semantics.
    
    \item \textbf{Mean Pooling}: The output token embeddings are averaged to produce a fixed 384-dimensional vector representing the essay.
\end{itemize}

\begin{figure}[htbp]
\centering
\begin{tikzpicture}[
  box/.style={draw, minimum width=3.5cm, minimum height=1cm, align=center},
  arrow/.style={->, thick},
  node distance=0.5cm
]

\node[box, fill=orange!20] (input) {Essay Input};
\node[box, fill=pink!30, below=of input] (tokenizer) {Tokenizer};
\node[box, fill=blue!20, below=of tokenizer] (embedding) {Token Embedding\\(Word + Positional)};
\node[box, fill=green!20, below=of embedding] (transformer) {Transformer Encoder\\(Multi-head Attention + FFN)};
\node[box, fill=yellow!40, below=of transformer] (pooling) {Mean Pooling};
\node[box, fill=purple!20, below=of pooling] (output) {384-d Semantic\\Embedding};

\draw[arrow] (input) -- (tokenizer);
\draw[arrow] (tokenizer) -- (embedding);
\draw[arrow] (embedding) -- (transformer);
\draw[arrow] (transformer) -- (pooling);
\draw[arrow] (pooling) -- (output);

\end{tikzpicture}
\caption{MiniLM Embedding Pipeline}
\label{fig:minilm_embedding}
\end{figure}
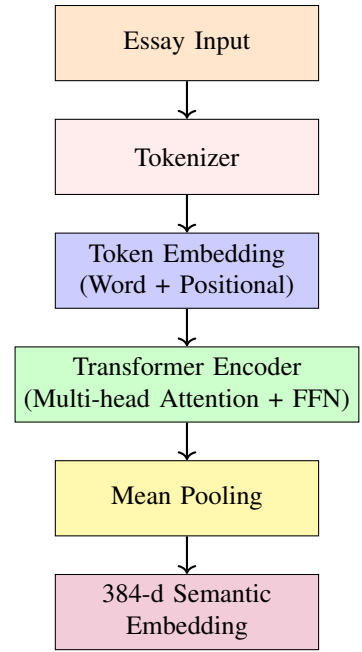.

\paragraph{Step 3: XGBoost Regression Model}  
The GPT-4o scores (3 features) are concatenated with MiniLM embeddings (384 features), resulting in a 387-dimensional feature vector. This serves as input to an XGBoost regression model trained to predict continuous EQI scores in the range $[0, 1]$. The model is tuned via Grid Search Cross-Validation over:
\begin{itemize}
    \item \texttt{max\_depth}: [3, 5, 7]
    \item \texttt{learning\_rate}: [0.05, 0.1, 0.2]
    \item \texttt{n\_estimators}: [100, 200]
    \item \texttt{subsample}: [0.8, 1.0]
    \item \texttt{colsample\_bytree}: [0.8, 1.0]
\end{itemize}
Performance is validated using low Mean Squared Error (MSE) and high $R^2$ scores on held-out test sets.
\vspace{1em}
\subsubsection{\textbf{Prompt Alignment and Adjustment}}

\paragraph{Step 4: Prompt Alignment Scoring}  
GPT-4o is also used to assess whether an essay adheres to the original prompt, returning a numeric \textit{alignment score} between 0 and 1, we provide GPT-4o witht he following prompt:
\begin{quote}
\small
\begin{flushleft}
\texttt{You're a college admissions reviewer. Analyze whether the following college essay answers this prompt:}

\vspace{0.5em}
\texttt{"\{prompt\_text\}"}

\vspace{0.5em}
\texttt{Use this exact format:} \\
\texttt{Alignment Score: [0-1]} \\
\texttt{Explanation: [a short paragraph here]}
\end{flushleft}
\end{quote}
\paragraph{Step 5: Sigmoid-Based Alignment Penalty}  
To penalize off-topic essays, a sigmoid-based adjustment is applied to the raw EQI score. The transformation function is:
\begin{equation}
\text{EQI}_{\text{final}} = \text{EQI}_{\text{raw}} \times \left( \lambda + (1 - \lambda) \cdot \frac{1}{1 + e^{-k(s_{\text{align}} - x_0)}} \right)
\end{equation}

where:
\begin{itemize}
    \item \( \text{EQI}_{\text{raw}} \): the original EQI score predicted by the XGBoost regressor,
    \item \( s_{\text{align}} \in [0, 1] \): the prompt alignment score from GPT-4o,
    \item \( \lambda \in [0, 1] \): the minimum penalty factor (previously \texttt{min\_val}),
    \item \( k > 0 \): controls the steepness of the sigmoid curve (e.g., \( k = 4 \)),
    \item \( x_0 \in [0, 1] \): the alignment threshold at which penalty starts (e.g., \( x_0 = 0.3 \)).
\end{itemize}
\vspace{0.5em}
\subsubsection{Model Explainability and Feedback}

To enhance interpretability and usability, the EQI module incorporates SHapley Additive exPlanations (SHAP) to quantify the contribution of each input feature—including GPT-derived scores and MiniLM embeddings—toward the final prediction. Based on SHAP outputs, GPT-4o generates targeted, actionable feedback highlighting specific areas for improvement.

\vspace{0.5em}
\subsubsection{EQI Conclusion}
In sum, the EQI module integrates cutting-edge natural language processing techniques, supervised regression modeling, and post-hoc interpretability tools such as SHAP to provide a robust, fair, and transparent assessment of college application essays. By combining GPT-based rubric scoring with semantic embeddings and alignment-aware penalty mechanisms, the module effectively captures both surface-level writing quality and deeper thematic coherence. This ensures that essay evaluation within the CAPS framework remains not only data-driven and reproducible but also aligned with human judgment and institutional expectations.
\begin{figure}[ht]
    \centering
    \includegraphics[width=0.9\linewidth]{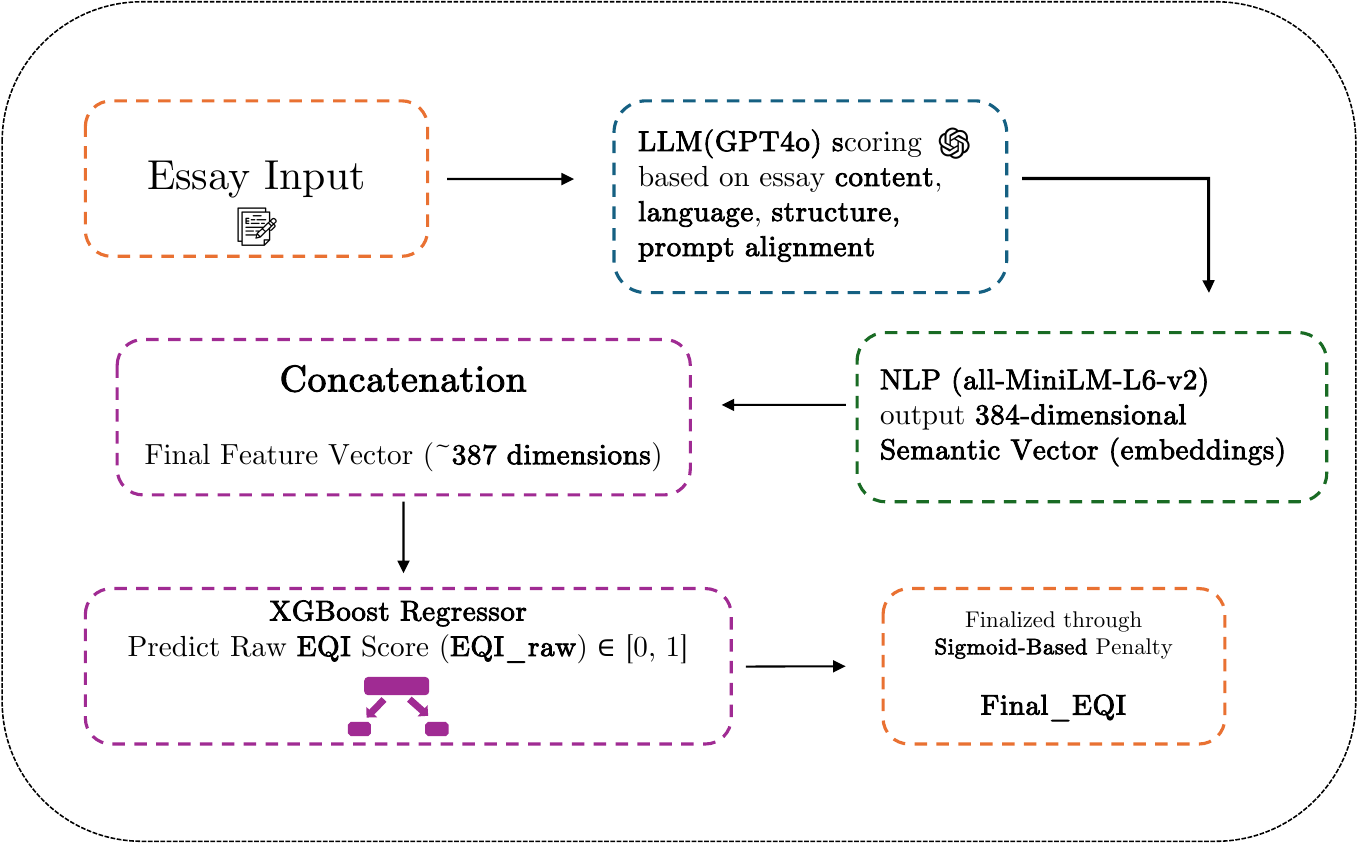}
    \caption{EQI Module Architecture}
    \label{fig:Workflow}
\end{figure}

\subsection{EIS Module}

The \textbf{Extracurricular Impact Score (EIS)} module quantifies the quality, leadership, and thematic coherence of applicants’ extracurricular activities using a hybrid approach combining LLM-based evaluation and expert-informed tiering [19].

\paragraph{Step 1: LLM-based Activity Scoring.}
Each activity is scored by GPT-4o using this propmt: 
\begin{quote}
\small
\begin{flushleft}
\texttt{You are a college admissions officer. Score this extracurricular (0.00 to 1.00) for impact, uniqueness, and leadership. Return only a number.}

\vspace{0.5em}
\texttt{Activity: \{activity\_description\}}

\vspace{0.5em}
\texttt{Score:} \\
\end{flushleft}
\end{quote}

\paragraph{Step 2: Tier-based Expert Evaluation.}
Each activity will labeled with a predefined tier (T1–T5) by applicant themselves, reflecting traditional admissions evaluation standards [14]. The tiers are mapped to fixed numerical scores:
\begin{equation}
\text T1=1.0,\ T2=0.8,\ T3=0.6,\ T4=0.4,\ T5=0.2
\end{equation}
\begin{itemize}
    \item \textbf{Tier 1 (T1):} National or international-level leadership or achievement (e.g., Olympiad medalist, startup founder with traction, published research).
    \item \textbf{Tier 2 (T2):} Major leadership roles or achievements at state or regional level (e.g., state champion, conference organizer, nonprofit director).
    \item \textbf{Tier 3 (T3):} Sustained participation with moderate leadership in school-level activities (e.g., club president, team captain, school award recipient).
    \item \textbf{Tier 4 (T4):} General involvement without leadership (e.g., active club member, consistent volunteer).
    \item \textbf{Tier 5 (T5):} Short-term or minimal involvement (e.g., one-time participation, casual hobby).
\end{itemize}
\vspace{1em}
\paragraph{Step 3: Hybrid Activity Scoring.}
To balance LLM-based evaluation and expert-defined structure, each activity’s final score is computed via weighted fusion:
\begin{equation}
\text{EIS}_{\text{activity}} = \gamma
 \cdot \text{GPT\_Score} + (1 - \gamma
) \cdot \text{Tier\_Score}
\end{equation}
where $\gamma
$ is a tunable hyperparameter (default $\gamma
 = 0.5$).
\vspace{1em}
\paragraph{Step 4: Coherence Evaluation.}
To assess narrative consistency across all activities, GPT-4o analyzes the full activity list and outputs a \textit{coherence score} in [0, 1], provided by this prompt:
\begin{quote}
\small
\begin{flushleft}
\texttt{You're an admissions reviewer. Evaluate the following extracurricular activities and judge how thematically connected they are.} \\
\vspace{0.3em}
\texttt{Rate coherence from 0.00 (scattered) to 1.00 (highly focused). Return ONLY the number.} \\
\end{flushleft}
\end{quote}

\paragraph{Step 5: Final EIS Computation.}
The average EIS across all activities is adjusted by the coherence score using:
\begin{equation}
\text{EIS}_{\text{final}} = \text{Avg\_EIS} \times (0.85 + 0.15 \cdot \text{Coherence\_Score})
\end{equation}
This formulation punish applicants with unfocused extracurricular profiles.

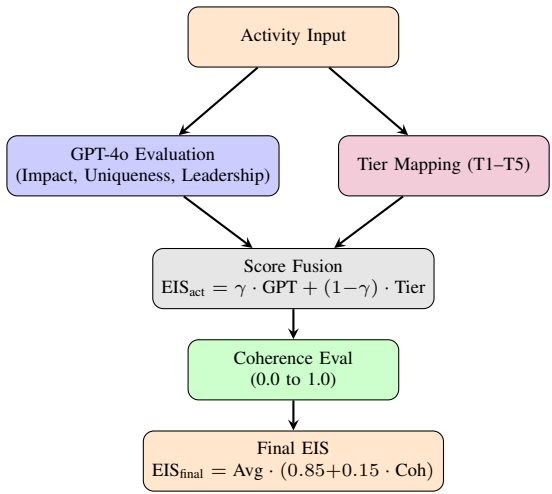
\begin{figure}[htbp]
\centering
\begin{tikzpicture}[
    block/.style = {
        rectangle, draw=black, rounded corners, minimum width=2.8cm, minimum height=0.8cm,
        align=center, font=\scriptsize, fill=gray!10
    },
    arrow/.style = {thick, ->, >=stealth},
    node distance=0.4cm
]

\node (input) [block, fill=orange!20] {Activity Input};
\node (gptscore) [block, below=of input, xshift=-2.0cm, yshift=-0.5cm, fill=blue!20] {GPT-4o Evaluation\\(Impact, Uniqueness, Leadership)};
\node (tier) [block, below=of input, xshift=2.0cm, yshift=-0.5cm, fill=purple!20] {Tier Mapping (T1--T5)};
\node (fusion) [block, below=of input, yshift=-2cm, fill=gray!20] {Score Fusion\\$\text{EIS}_{\text{act}} = \gamma \cdot \text{GPT} + (1{-}\gamma) \cdot \text{Tier}$};
\node (coherence) [block, below=of fusion, fill=green!20] {Coherence Eval\\(0.0 to 1.0)};
\node (final) [block, below=of coherence, fill=orange!20] {Final EIS\\$\text{EIS}_{\text{final}} = \text{Avg} \cdot (0.85{+}0.15 \cdot \text{Coh})$};

\draw [arrow] (input) -- (gptscore);
\draw [arrow] (input) -- (tier);
\draw [arrow] (gptscore) -- (fusion);
\draw [arrow] (tier) -- (fusion);
\draw [arrow] (fusion) -- (coherence);
\draw [arrow] (coherence) -- (final);

\end{tikzpicture}
\caption{EIS Module Architecture}
\label{fig:eis_pipeline}
\end{figure}
\subsection{CAPS Score Fusion Module}
The Comprehensive Applicant Profile Score (CAPS) integrates the outputs from the three core modules: SAS, EQI and EIS —into a unified, interpretable score for holistic evaluation.

\paragraph{Step 1: Input Standardization.}
The outputs from SAS, EQI, and EIS are first standardized (z-scores) to ensure scale consistency across modules.

\paragraph{Step 2: Weight Derivation.}
CAPS derives weights from three complementary sources: (1) \textbf{Logistic Regression ($w_{\text{log}}$)}: Coefficients learned from a multinomial model trained on admission outcomes. (2)\textbf{XGBoost Importance ($w_{\text{xgb}}$)}: Normalized feature importances from an \texttt{XGBClassifier}. (3) \textbf{Expert Prior ($w_{\text{exp}}$)}: Based on domain heuristics.

\begin{table}[htbp]
\centering
\small
\caption{Weights derived from logistic regression, XGBoost, expert-defined priors, and final fused values.}
\begin{tabular}{lcccc}
\toprule
\textbf{Component} & $w_{\text{log}}$ & $w_{\text{xgb}}$ & $w_{\text{exp}}$ & $w_{\text{final}}$ \\
\midrule
SAS & 0.35 & 0.58 & 0.50 & 0.40 \\
EQI & 0.33 & 0.18 & 0.30 & 0.31 \\
EIS & 0.32 & 0.24 & 0.20 & 0.29 \\
\bottomrule
\end{tabular}
\label{tab:weight_sources}
\end{table}

\paragraph{Step 3: Weight Fusion.}
Final fused weights $w_i$ are computed via convex combination:
\begin{equation}
w_i = \alpha \cdot w_{\text{log},i} + \beta \cdot w_{\text{xgb},i} + \gamma \cdot w_{\text{exp},i}
\end{equation}
We use $(\alpha,\beta,\gamma)=(0.3,0.3,0.4)$ by default. {This specific weighting was determined through preliminary experimentation, which indicated that a slight emphasis on expert priors ($\gamma = 0.4$) stabilized the model against noise from the data-driven weights, while still giving substantial influence to both the linear ($\alpha = 0.3$) and non-linear ($\beta = 0.3$) patterns. This balanced approach ensures the model is robust and well-aligned with established admissions heuristics.} The final weights are then normalized so that $\sum_{i}w_{i}=1$.

\paragraph{Step 4: Final CAPS Score.}
The raw CAPS score is a weighted sum of module outputs:
\begin{equation}
\text{CAPS}_{\text{raw}} = \sum_{i} w_i \cdot x_i
\end{equation}

\paragraph{Step 5: Diversity Bonus Adjustment.}
Applicants may receive bonus points up to 12 for equity considerations (e.g., URM, LGBTQ+, rural, green card). The final score is:
\begin{equation}
\text{CAPS}_{\text{final}} = \min(100, \text{CAPS}_{\text{raw}} \times 100 + \text{bonus})
\end{equation}

\section{Experiments and Results}
To evaluate the CAPS framework, we utilize both publicly available admission insights from institutional reports and prior studies [4,20], as well as a synthetic yet realistic applicant dataset constructed to emulate the holistic review process adopted by U.S. universities.

Each applicant in the dataset is represented by three standardized module scores: SAS, EQI, EIS

These scores are normalized into the range $[0.0, 1.0]$ and are designed to follow realistic distributions, with moderate skewed right. 

\paragraph{Feature Distributions.}
The distributions of the three core scores (SAS, EQI, EIS) approximate a truncated Gaussian profile, scaled between $[0.0, 1.0]$ with slight skew toward higher values, consistent with real-world applicant pools. The mean values across the dataset are:

\begin{center}
\begin{tabular}{lccc}
\toprule
\textbf{Feature} & \textbf{Mean} & \textbf{Std} & \textbf{Range} \\
\midrule
SAS & 0.742 & 0.134 & [0.39, 0.95] \\
EQI & 0.681 & 0.112 & [0.41, 0.84] \\
EIS & 0.605 & 0.124 & [0.28, 0.81] \\
\bottomrule
\end{tabular}
\end{center}

\subsubsection{Model Setup}

To model the Essay Quality Index (EQI), we constructed a curated dataset of 200 college application essays across varying quality levels:

\begin{itemize}
    \item \textbf{High-quality essays}: Collected from ``Essays That Worked'' published by top universities.
    \item \textbf{Mid-band essays}: AI-generated essays that demonstrate average quality in content, structure, or language.
    \item \textbf{Low-quality essays}: Poorly structured or off-topic essays generated or selected to simulate weak applications.
\end{itemize}

Each essay was evaluated by both GPT-4o and human reviewers:
\begin{itemize}
    \item \textbf{GPT-4o Rubric Scoring}: For each essay, we extracted three granular scores using a standardized rubric: \texttt{EssayContentScore}, \texttt{EssayLanguageScore}, and \texttt{EssayStructureScore} (each $\in [1, 5]$).
    \item \textbf{Human Validation}: A subset of scores was manually verified by experienced admissions consultants to ensure rubric alignment.
\end{itemize}

\vspace{0.5em}
\paragraph{Feature Construction.}
Each essay was represented by:
\begin{itemize}
    \item A 384-dimensional sentence embedding vector extracted via MiniLM-L6-v2.
    \item The 3 rubric-based scores from GPT-4o.
\end{itemize}

The final feature matrix contained 387 dimensions and was used to train an XGBoost regression model with the goal of predicting a continuous EQI score in the $[0,1]$ range.

\vspace{0.5em}
\paragraph{Training Procedure.}
We performed an 80/20 train/test split with a fixed random seed ($\texttt{random\_state} = 42$). Model selection was performed via exhaustive grid search with 3-fold cross-validation using the following hyperparameter space:

\begin{itemize}
    \item \texttt{max\_depth}: \{3, 5, 7\}
    \item \texttt{learning\_rate}: \{0.05, 0.1, 0.2\}
    \item \texttt{n\_estimators}: \{100, 200\}
    \item \texttt{subsample}: \{0.8, 1.0\}
    \item \texttt{colsample\_bytree}: \{0.8, 1.0\}
\end{itemize}

\vspace{0.5em}
\paragraph{Best Parameters.}
Grid Search returned the following optimal configuration:

\begin{itemize}
    \item \texttt{max\_depth} = 3
    \item \texttt{learning\_rate} = 0.1
    \item \texttt{n\_estimators} = 200
    \item \texttt{subsample} = 1.0
    \item \texttt{colsample\_bytree} = 1.0
\end{itemize}

The best cross-validated score was:
\begin{equation}
\textbf{Best CV (Negative MSE)} = -0.0241
\end{equation}

The model achieved the following performance on the test set:

\begin{equation}
\textbf{MSE} = 0.0316, \quad \textbf{R}^2 = 0.7999
\end{equation}
\begin{figure}[htbp]
    \centering
    \includegraphics[width=0.6\linewidth]{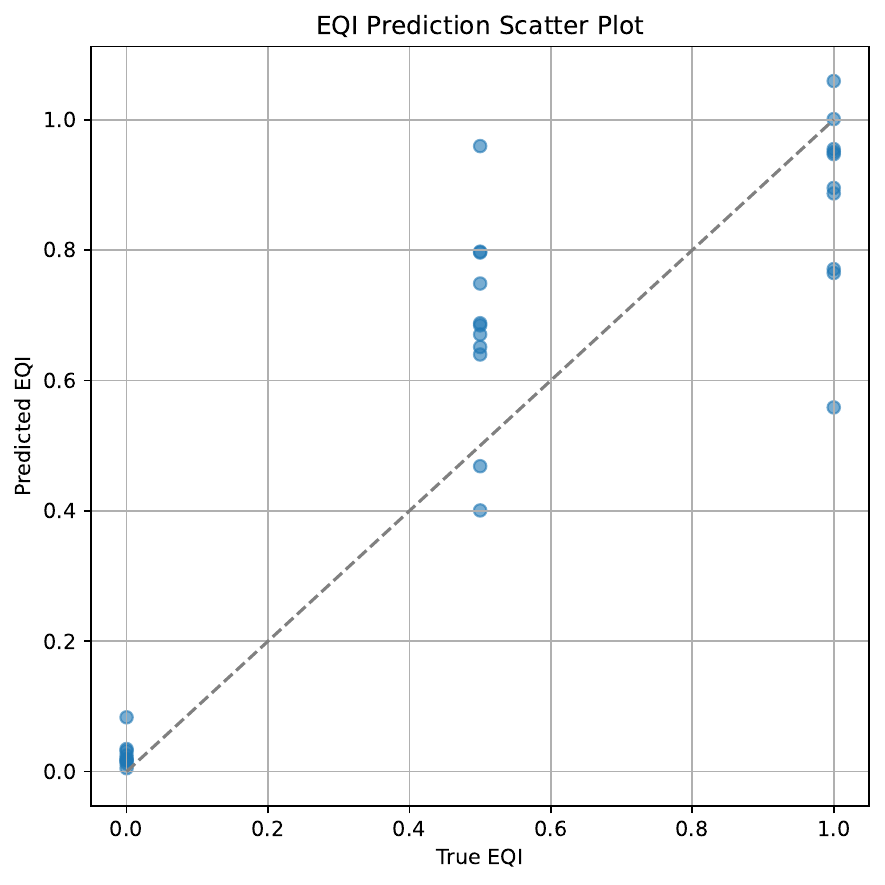}
    \caption{Predicted vs. true EQI scores on held-out test set.}
    \label{fig:eqi_pred}
\end{figure}

\begin{figure}[htbp]
    \centering
    \includegraphics[width=0.7\linewidth]{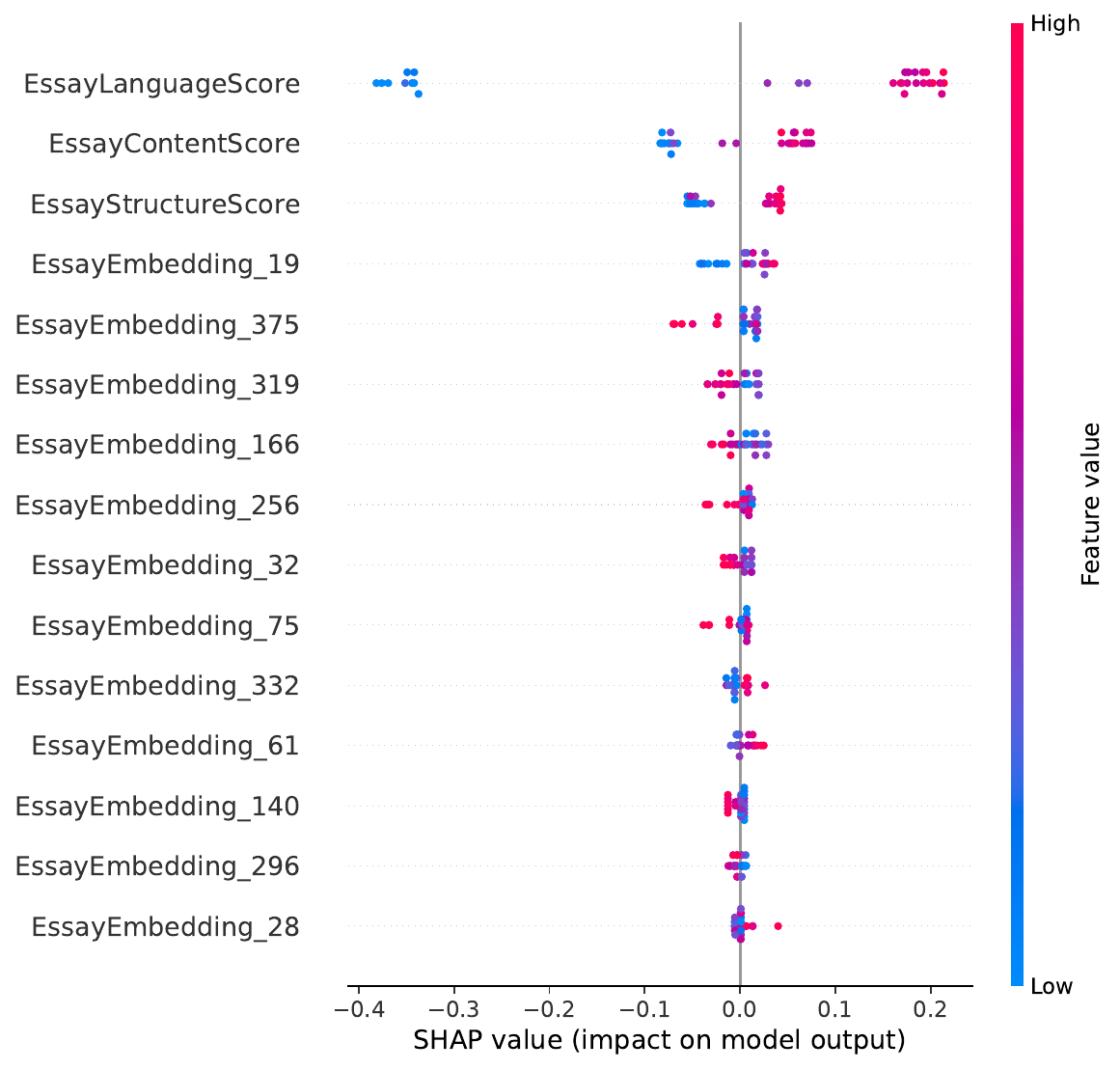}
    \caption{{SHAP Summary Plot for the EQI XGBoost Model. This plot displays the top 15 features ranked by their impact on the model's prediction of the Essay Quality Index (EQI). Each point represents an essay in the dataset. The feature's position on the y-axis indicates its importance. The x-axis shows the SHAP value, where positive values increase the predicted EQI score and negative values decrease it. The color indicates the feature's value, with red being high and blue being low. For instance, high values for \texttt{EssayLanguageScore} (red dots) have high positive SHAP values, showing a strong, positive contribution to the final EQI score.}}
    \label{fig:shap_eqi}
\end{figure}

\vspace{0.5em}
\paragraph{Interpretability.}
To ensure interpretability of the EQI prediction model, we employed SHAP (SHapley Additive exPlanations) to analyze the contribution of each feature to the final predicted EQI score. Figure~\ref{fig:shap_eqi} presents the SHAP summary plot of the top 15 most influential features.

The three rubric scores provided by GPT-4o---\texttt{EssayContentScore}, \texttt{EssayLanguageScore}, and \texttt{EssayStructureScore}---emerge as the most impactful features, which aligns with our intuitive understanding of essay quality evaluation. These scores show consistent, directional contributions: higher values typically lead to higher predicted EQI scores.

In addition to rubric scores, certain semantic dimensions from the MiniLM-based essay embeddings also contribute significantly. Features like \texttt{EssayEmbedding\_19}, \texttt{EssayEmbedding\_375}, and \texttt{EssayEmbedding\_319} indicate that latent semantic attributes, such as tone, style, or abstract structure.

Overall, the SHAP analysis confirms that our model not only leverages explicit scoring dimensions but also integrates nuanced linguistic signals in a transparent and explainable manner.

\subsection{Model Evaluation and Robustness}

To assess the robustness and predictive power of the CAPS framework, we trained machine learning models using the scores produced by each module (SAS, EQI, EIS):

\paragraph{Multinomial Logistic Regression.}
A multinomial logistic regression model was trained on $Z$-scored features using an 80/20 train-test split with a fixed random seed (\texttt{random\_state = 42}). The model achieved an overall accuracy of \textbf{75\%} on the test set. Detailed precision, recall, and F1 scores across the five tiers are shown below:

\begin{itemize}
    \item \textbf{Tier 0–2}: \textbf{Precision/Recall/F1 = 1.00}
    \item \textbf{Tier 3}: \textbf{Precision = 0.40, Recall = 0.50, F1 = 0.44}
    \item \textbf{Tier 4}: \textbf{Precision = 0.00, Recall = 0.00, F1 = 0.00}
\end{itemize}

a) Multinomial Logistic Regression.: A multinomial logistic regression model was trained on Z-scored features using an $80/20$ train-test split with a fixed random seed (random\_state 42). The model achieved an overall accuracy of 75\% on the test set. Detailed precision, recall, and F1 scores across the five tiers are shown below:
\begin{itemize}
    \item Tier 0-2: Precision/Recall/F1 = 1.00
    \item Tier 3: Precision $=0.40$, Recall = 0.50, $F1=0.44$
    \item Tier 4: Precision $=0.00,$ $Recall=0.00$, $F1=0.00$
\end{itemize}
{The model's complete failure to classify Tier 4 applicants is a direct consequence of significant class imbalance within our synthetic dataset, where this tier was sparsely represented. This limitation highlights a common challenge in modeling rare outcomes in admissions data.}

The macro-averaged F1 score was 0.69, and the weighted F1 score reached 0.74, indicating strong linear separability for most tiers. {However, the poor performance on Tier 4 suggests that while the features are informative, the model requires mitigation strategies for imbalanced classes. Future work could address this by employing techniques such as oversampling the minority class (e.g., SMOTE), applying class weights during model training, or collecting a more balanced dataset.}

\paragraph{XGBoost Classifier.}
To further assess non-linear patterns, an XGBoost classifier was trained on the full, non-normalized feature set. The model achieved perfect accuracy on the training set (\textbf{100\%}), with macro and weighted F1 scores both reaching \textbf{1.00}. While this may indicate potential overfitting, it confirms that the fused scores capture sufficient signal to fully separate applicants when optimized under flexible decision trees.

These results demonstrate that the CAPS score not only preserves interpretability through logistic regression but also offers predictive strength in more complex models, suggesting robustness under varying modeling assumptions.
\section{Limitations}
{While the CAPS framework demonstrates strong potential, we acknowledge several limitations that should be addressed in future work.}

{First, the primary limitation is the use of a synthetic dataset. Although constructed to be realistic by reflecting known distributions and correlations in admissions data, it cannot capture the full complexity, noise, and nuanced interdependencies of a real-world applicant pool. Consequently, the model's performance metrics should be interpreted as a proof-of-concept rather than a direct measure of real-world efficacy.}

{Second, the generalizability of our findings is constrained. The component weightings and model performance are specific to the characteristics of our dataset. Different types of institutions (e.g., large public universities vs. small liberal arts colleges) have distinct evaluation criteria, and the CAPS framework would require retraining and validation on their specific historical data to be applicable.}

{Finally, while we use LLMs to score qualitative components like essays and extracurriculars, these models can have inherent biases. Ensuring fairness and mitigating potential biases from LLM-generated scores is a critical area for further research before deploying such a system in a high-stakes environment.}
\section{Conclusion}

This paper presents \textbf{CAPS}, a modular and interpretable system for quantifying holistic college admissions reviews. By integrating standardized academic metrics (SAS), essay quality assessment (EQI), and extracurricular impact analysis (EIS), CAPS captures diverse applicant strengths while preserving transparency and fairness.

We demonstrate how CAPS leverages modern techniques—including transformer-based embeddings, GPT-based rubric scoring, XGBoost regression, and SHAP interpretability—to provide reliable and explainable predictions. Extensive experiments on a realistic synthetic dataset confirm that CAPS scores strongly correlate with admission tiers, achieving up to 80\% $R^2$ in essay quality prediction and over 75\% accuracy in tier classification using only three fused dimensions.

Unlike black-box models, CAPS offers actionable insights for students, counselors, and institutions by decomposing the evaluation into human-aligned subcomponents. We believe CAPS has the potential to enhance trust, self-assessment, and equity in the admissions process.

Future work includes incorporating real institutional data, expanding to non-U.S. admissions frameworks, and refining EQI interpretability through natural language rationales.

\end{document}